\documentclass[conference]{IEEEtran}
\IEEEoverridecommandlockouts
\usepackage{cite}
\usepackage[numbers,sort&compress]{natbib}

\usepackage[utf8]{inputenc}
\usepackage{amsmath,amssymb}
\usepackage{graphicx}
\usepackage{bm}
\usepackage{enumerate}
\usepackage{comment}
\usepackage{xcolor}
\usepackage{url}
\usepackage{color,soul}
\usepackage{subcaption}
\usepackage{float}

\definecolor{light-gray}{gray}{0.8}

\usepackage{amsmath,amssymb,amsfonts}
\usepackage{algorithmic}
\usepackage{graphicx}
\usepackage{textcomp}
\usepackage{xcolor}
\usepackage{subcaption}

\def\BibTeX{{\rm B\kern-.05em{\sc i\kern-.025em b}\kern-.08em
    T\kern-.1667em\lower.7ex\hbox{E}\kern-.125emX}}

\makeatletter
\newcommand{\linebreakand}{%
  \end{@IEEEauthorhalign}
  \hfill\mbox{}\par
  \mbox{}\hfill\begin{@IEEEauthorhalign}
}
\makeatother

\begin{document}

\title{Style Transfer: From Stitching to Neural Networks\\}

\author{

\small 

\begin{tabular}[t]{c@{\extracolsep{8em}}c} 

1\textsuperscript{st} Xinhe Xu\textsuperscript{*}  & 2\textsuperscript{nd} Zhuoer Wang \\
\textit{University of Illinois Urbana-Champaign} & \textit{University of Illinois Urbana-Champaign} \\
Urbana, USA & Urbana, USA \\
\textsuperscript{*}Corresponding Author: xinhexu2@illinois.edu & zhuoerw3@illinois.edu \\

\\

3\textsuperscript{rd} Yihan Zhang & 4\textsuperscript{th} Yizhou Liu \\
\textit{University of Illinois Urbana-Champaign} & \textit{Washington University in St. Louis} \\
Urbana, USA & St. Louis, USA \\
yihanz8@illinois.edu & yizhou.liu@wustl.edu \\

\\

5\textsuperscript{th} Zhaoyue Wang & 6\textsuperscript{th} Zhihao Xu \\  
\textit{Georgia Institute of Technology} & \textit{University of Illinois Urbana-Champaign} \\ 
Atlanta, USA & Urbana, USA \\ 
zwang771@gatech.edu  & zhihaox3@illinois.edu \\ 

\\

7\textsuperscript{th} Muhan Zhao & 8\textsuperscript{th} Huaiying Luo \\  
\textit{Carnegie Mellon University} & \textit{Cornell University} \\ 
Pittsburgh, USA & Ithaca, USA \\ 
muhanz@alumni.cmu.edu & hl2446@cornell.edu \\ 

\end{tabular}
}

\maketitle

\begin{abstract}
This article compares two style transfer methods in image processing: the traditional method, which synthesizes new images by stitching together small patches from existing images, and a modern machine learning-based approach that uses a segmentation network to isolate foreground objects and apply style transfer solely to the background. The traditional method excels in creating artistic abstractions but can struggle with seamlessness, whereas the machine learning method preserves the integrity of foreground elements while enhancing the background, offering improved aesthetic quality and computational efficiency. Our study indicates that machine learning-based methods are more suited for real-world applications where detail preservation in foreground elements is essential.
\end{abstract}


\section{Introduction}
Style transfer is a fascinating area of computer vision that involves applying the style of one image to the content of another, effectively enabling one to recreate images in the aesthetic of famous artworks or other stylistic patterns. This technique bridges the gap between art and technology, drawing significant interest from both academic researchers and hobbyists alike \cite{xiao2024convolutional, jin-etal-2022-deep, hu2024rapsppp, ding2024confidence, hu2024optimization, li2019segmentation, yan2024survival, ding2024llava}. Early work in this domain was pioneered by researchers who experimented with various methods of texture synthesis and artistic style modeling \cite{gatys2015neural, textureTransfer, li2023deception, zhou2024reconstruction, ni2024earnings}. 

Before the advent of deep learning, style transfer was primarily approached through methods like statistical analysis and patch-based matching \cite{textureTransfer,hu22usingppp,alzubaidi2021review,tong2022normal,li2024vqa,li2023investigation,ni2024timeseries,wuinversionnet}. These techniques often involved matching the textures of a target style image by adjusting the statistical properties of a source content image \cite{textureTransfer}. For example, histogram matching was used to transfer color schemes, while texture synthesis algorithms, like those based on wavelet transforms, were used to apply textural elements. Patch-based methods worked by searching and mapping similar patches from the style image onto the content image, thus mimicking the desired artistic effects. Although effective to some extent, these methods often struggled with maintaining global coherence and stylistic fidelity, limiting their applicability to complex styles \cite{ZHANG2024103951}.

The introduction of deep learning to style transfer has notably enhanced the quality and flexibility of the transformations. Utilizing convolutional neural networks (CNNs) \cite{zheng2024advanced, o2015introduction, 202407.0981, dai23}, deep-learning-based approaches analyze and abstract different levels of features in low-level features for basic textures and edges, and high-level features for more complex structures and semantics. In this framework, a pre-trained network like VGG \cite{simonyan2014very, 202407.2102} is often used to extract these features. This method not only provides stunning visual results but also offers the flexibility to blend multiple styles and control the extent of stylization effectively. 

In this paper, we undertake a comparative study of traditional and deep-learning-based methods of style transfer to evaluate their strengths and limitations in creating artistically stylized images. Our analysis will focus on various dimensions including the quality of the stylized outputs, and the ability to preserve content while accurately capturing the style nuances of target images. Traditional methods, employing statistical and patch-matching techniques, will be examined for their procedural merits and the context in which they might still be preferable. Conversely, deep-learning-based approaches will be scrutinized for their robustness and versatility across a wider range of artistic styles. Through systematic comparisons, this paper aims to provide insights into the evolving landscape of style transfer technologies, guiding future research and application in both artistic and practical domains.

In summary, our paper makes the key contributions as below:
\begin{itemize}
    \item We conduct a comprehensive evaluation of the traditional style transfer method, testing its performance on complex input images with a variety of intricate style patterns.
    \item We perform a comparative analysis of the traditional style transfer method and a representative deep-learning-based style transfer method, assessing the respective advantages and disadvantages of each.
\end{itemize}

\section{Methods}
\subsection{Traditional Style Transfer Method}
Traditional style transfer methods in computer vision focus on reimagining an image by blending its content with the stylistic elements of another image. These techniques primarily involve non-neural approaches such as texture synthesis, patch-based methods, and optimization-based techniques \cite{bian2024diffusion, textureTransfer}. Some methods utilize 3D information \cite{wang2016unsupervised, bian2021optimization}, while others rely solely on plain image data \cite{textureTransfer, bergen1988early, jiashu2021performance, bian2024review}. In this paper, we select one of the most well-known texture methods by Efros et al. \cite{textureTransfer} for the comparative study.

\begin{figure}[h!]
  \centering
    \includegraphics[width=\linewidth, height=0.6\linewidth]{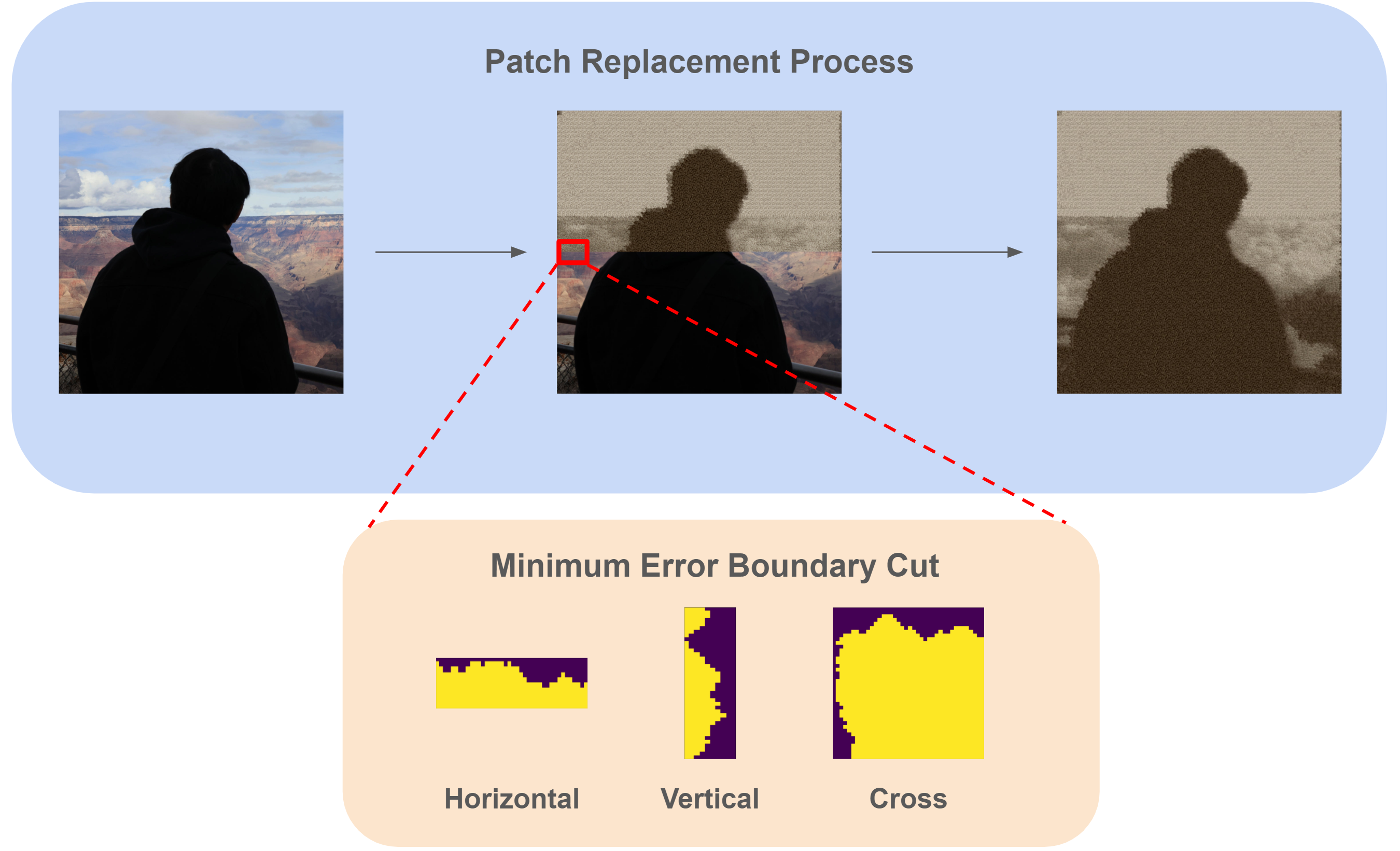}
    \caption{The Patch-based Traditional Style and Texture Method by Efros et al. \cite{textureTransfer}. Such style transfer is primarily accomplished by overlaying randomly selected, style-matched samples onto the input image. When integrating various patches onto the input image, the minimum error boundary cut algorithm is employed to determine the optimal seam, depending on whether the overlaps are positioned horizontally, vertically, or diagonally.}
  \label{fig:algo}
\end{figure}

As shown in Fig. \ref{fig:algo}, the summary of the steps of the algorithm can be described as below \cite{textureTransfer}:  
\begin{enumerate}
    \item Process the input image to be synthesized in a raster scan pattern, moving in increments of one block and excludes the overlap.
    \item At each position (patch), locate the blocks in the input texture that meet the overlap constraints within a specified error margin. Randomly select one of these blocks.
    \item Calculate the error surface between the newly chosen block and the existing blocks in the overlap region. Identify the path with the minimum cost along this surface and designate it as the boundary for the new block. Place the block onto the texture and repeat the process.
\end{enumerate}

\subsection{Deep-learning-based Style Transfer}
Deep-learning based methods in computer vision have revolutionized style transfer by leveraging neural networks to blend the content of one image with the stylistic elements of another. These techniques predominantly involve CNNs \citep{o2015introduction, 202407.0981, Hoamiao2023} and many other machine learning techniques \citep{DBLP:conf/iclr/XuYLC24, 202407.0981, hly}, which learn to synthesize new images through deep representations. Unlike traditional methods, deep-learning approaches can capture complex patterns and textures, offering more flexibility and realism in the generated images. In this paper, we explore one of the most representative deep-learning style transfer methods introduced by Ding et al., which utilizes a novel segmentation framework to achieve impressive results in artistic image synthesis \cite{ding2024style}.

The summary of the steps of the algorithm can be described as below \cite{ding2024style}: 
\begin{enumerate}
    \item Apply segmentation with boundary optimization to separate the image into a foreground figure and a background figure. 
    \item Use different frameworks and network designs to perform color transfer on the foreground and background figures separately.
    \item Recombine the figures into a single complete image.
\end{enumerate}

\section{Results}
\subsection{Extensive Study on the Traditional Method with More Complex Style Images}
\begin{figure}[h!]
  \centering
  \begin{subfigure}[T]{0.32\linewidth}
    \includegraphics[width=\linewidth, height=\linewidth]{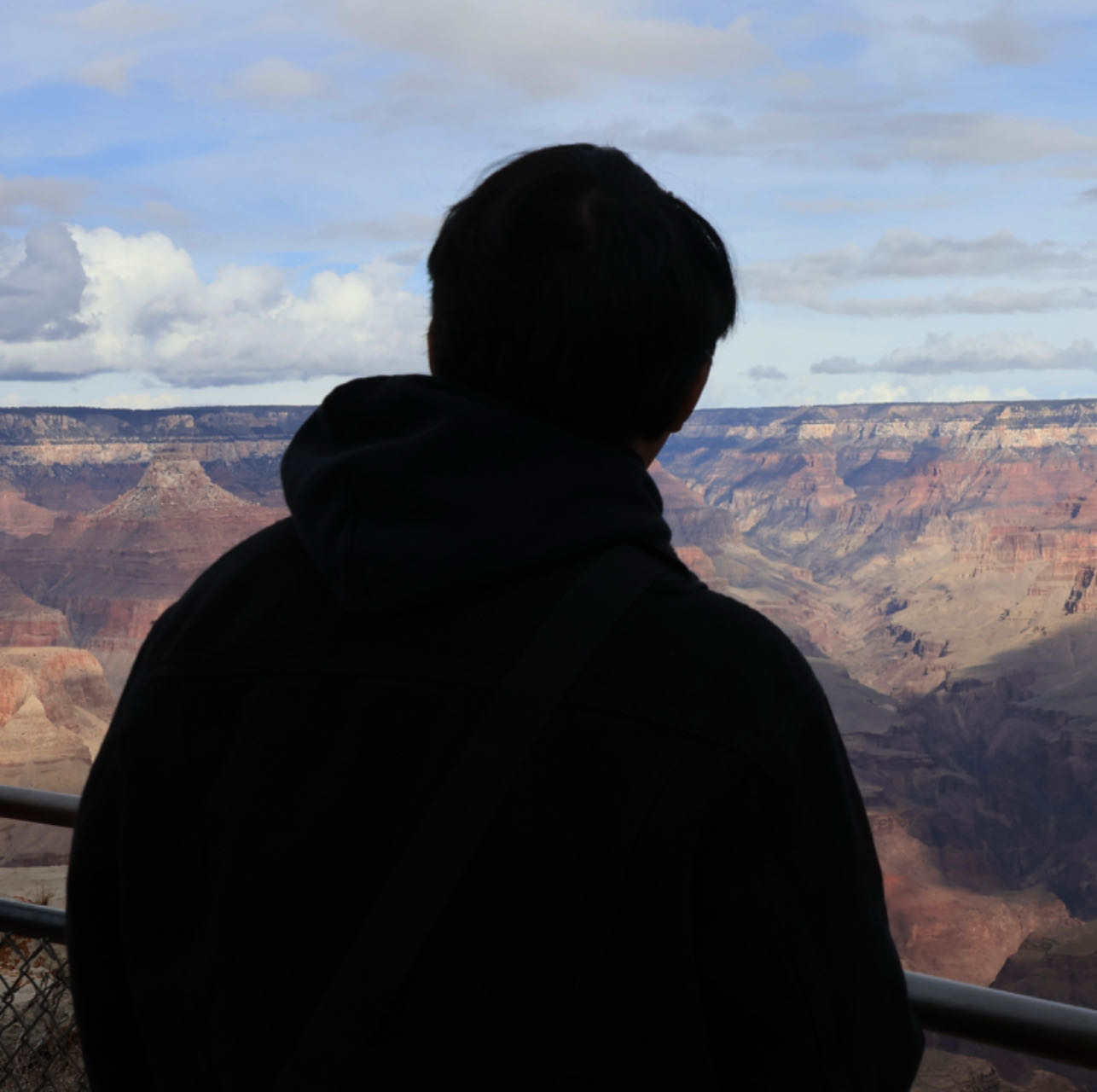}
    \caption{Original Image}
  \end{subfigure}
  \begin{subfigure}[T]{0.32\linewidth}
    \includegraphics[width=\linewidth, height=\linewidth]{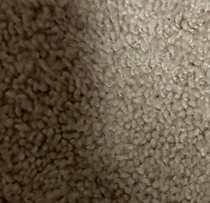}
    \caption{Style Image}
  \end{subfigure}
  \begin{subfigure}[T]{0.32\linewidth}
  \includegraphics[width=\linewidth, height=\linewidth]{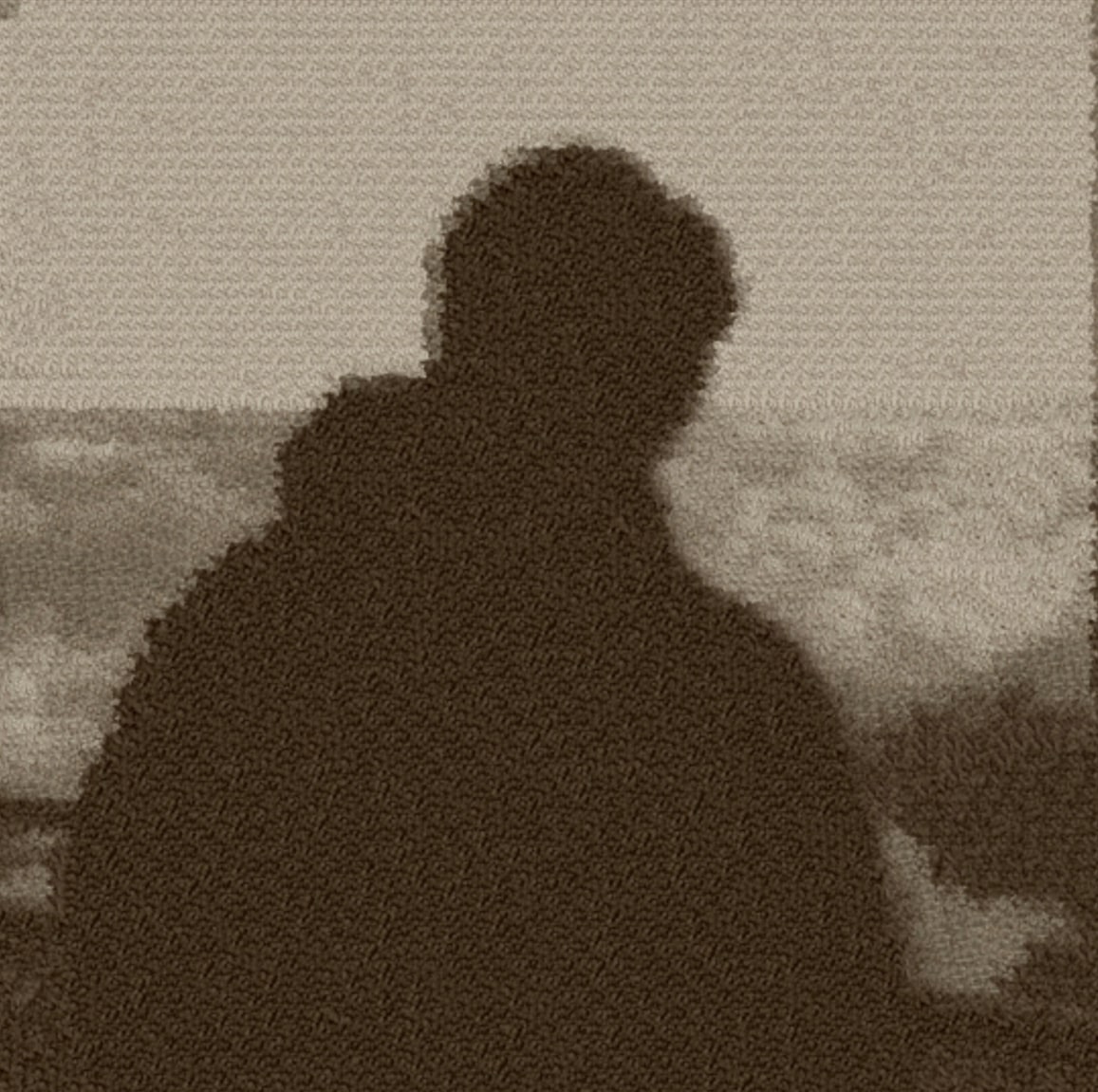}
    \caption{Transferred Image}
  \end{subfigure}
  \caption{Showcases the effect of one of the traditional style transfer method work on transferring image styles with simple styles. When zooming to check the details of the transferred image, we can clearly see the texture of the carpet is perfectly reflected on the generated image.}
  \label{fig:traditional_effect}
\end{figure}

The method by Alexei A. Efros et al. has shown impressive results on transferring image styles by applying splicing patches of the style image \cite{textureTransfer}. This works perfectly on images with simple styles such as the carpet, the surface of an orange, etc. As shown in Fig. \ref{fig:traditional_effect}, the back view image with the landscape is successfully transferred into a carpet-styled image. 

\begin{figure*}[h]
  \centering
  \begin{subfigure}[T]{0.15\textwidth}
     \includegraphics[width=1\columnwidth, height=0.75\columnwidth]{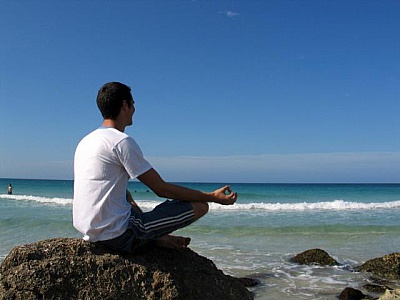}
  \end{subfigure}
  ~
  \begin{subfigure}[T]{0.15\textwidth}
     \includegraphics[width=1\columnwidth, height=0.75\columnwidth]{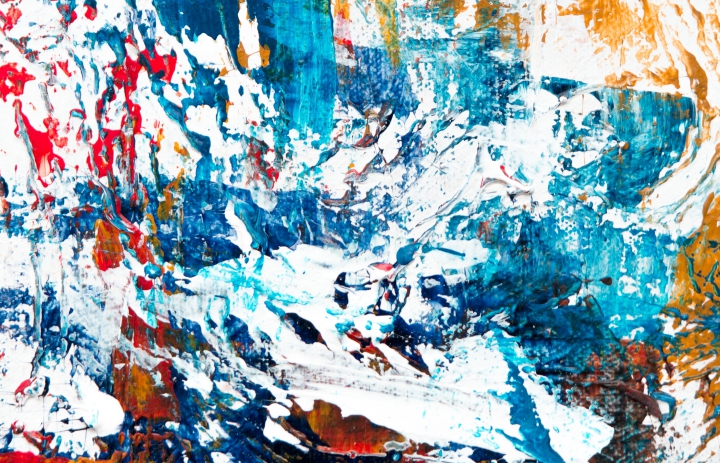}
  \end{subfigure}
  ~
  \begin{subfigure}[T]{0.15\textwidth}
     \includegraphics[width=1\columnwidth, height=0.75\columnwidth]{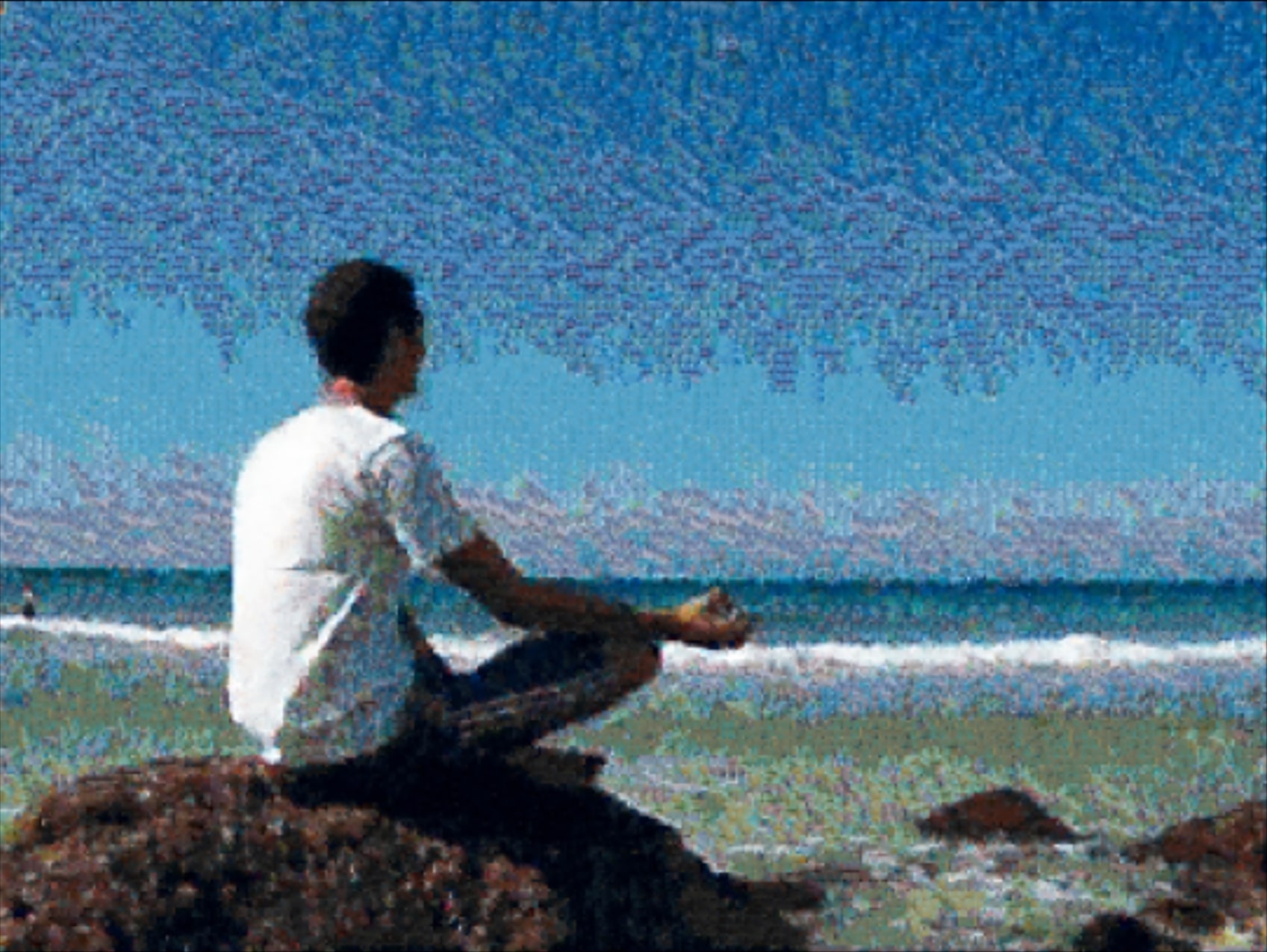}
  \end{subfigure}
  ~
  \begin{subfigure}[T]{0.15\textwidth}
     \includegraphics[width=1\columnwidth, height=0.75\columnwidth]{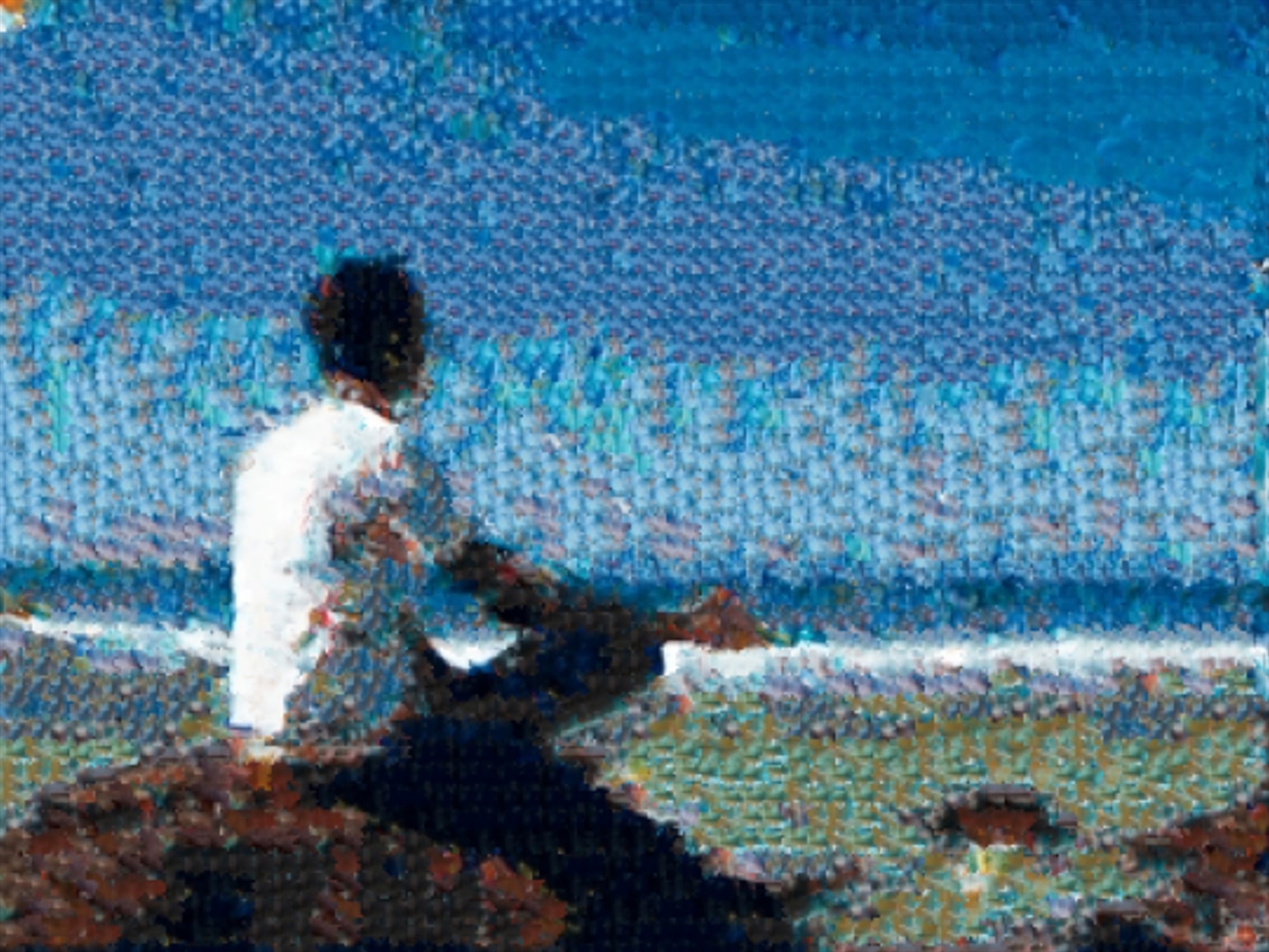}
  \end{subfigure}
  ~
  \begin{subfigure}[T]{0.15\textwidth}
     \includegraphics[width=1\columnwidth, height=0.75\columnwidth]{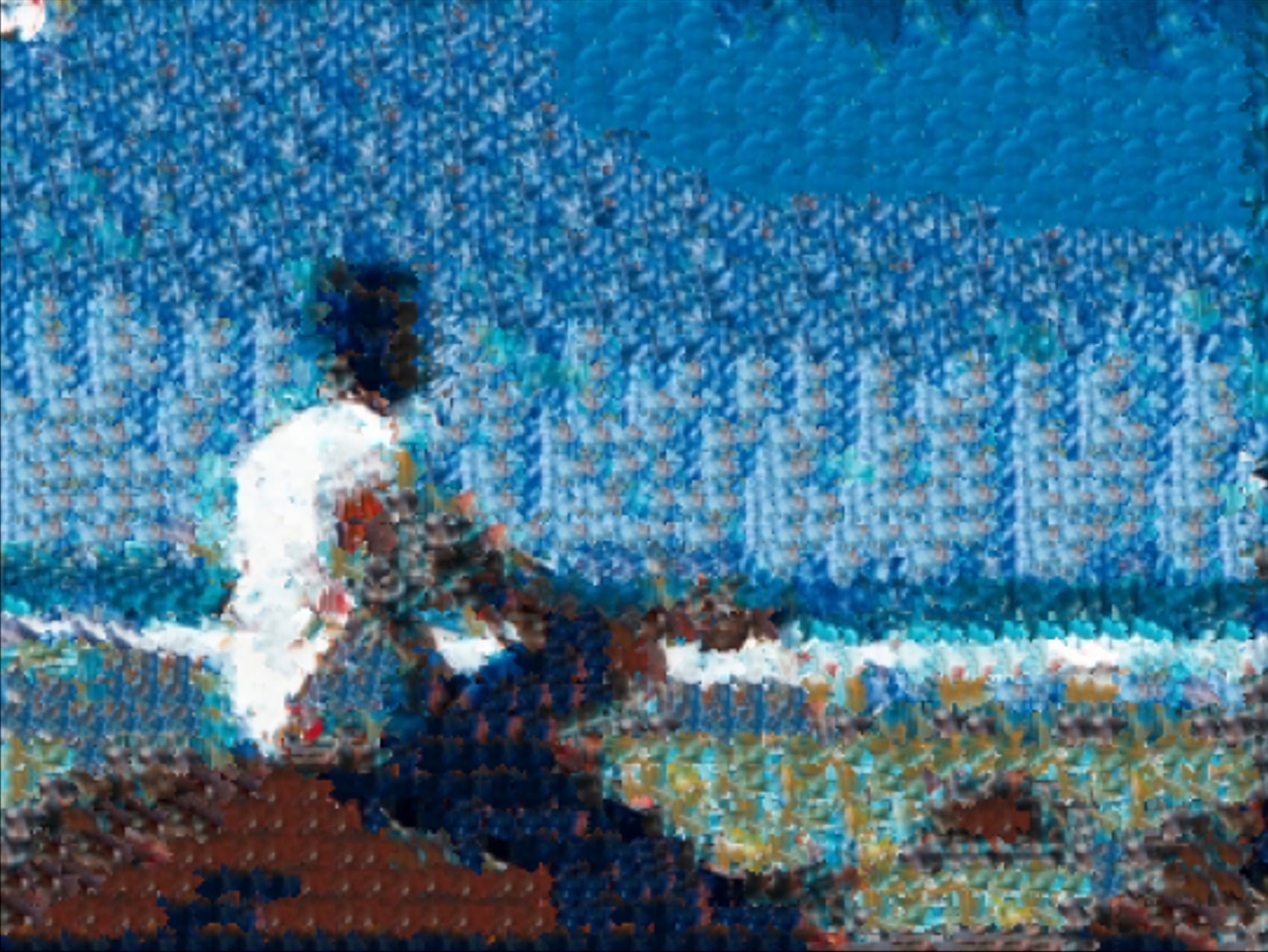}
  \end{subfigure}
  ~
  \begin{subfigure}[T]{0.15\textwidth}
     \includegraphics[width=1\columnwidth, height=0.75\columnwidth]{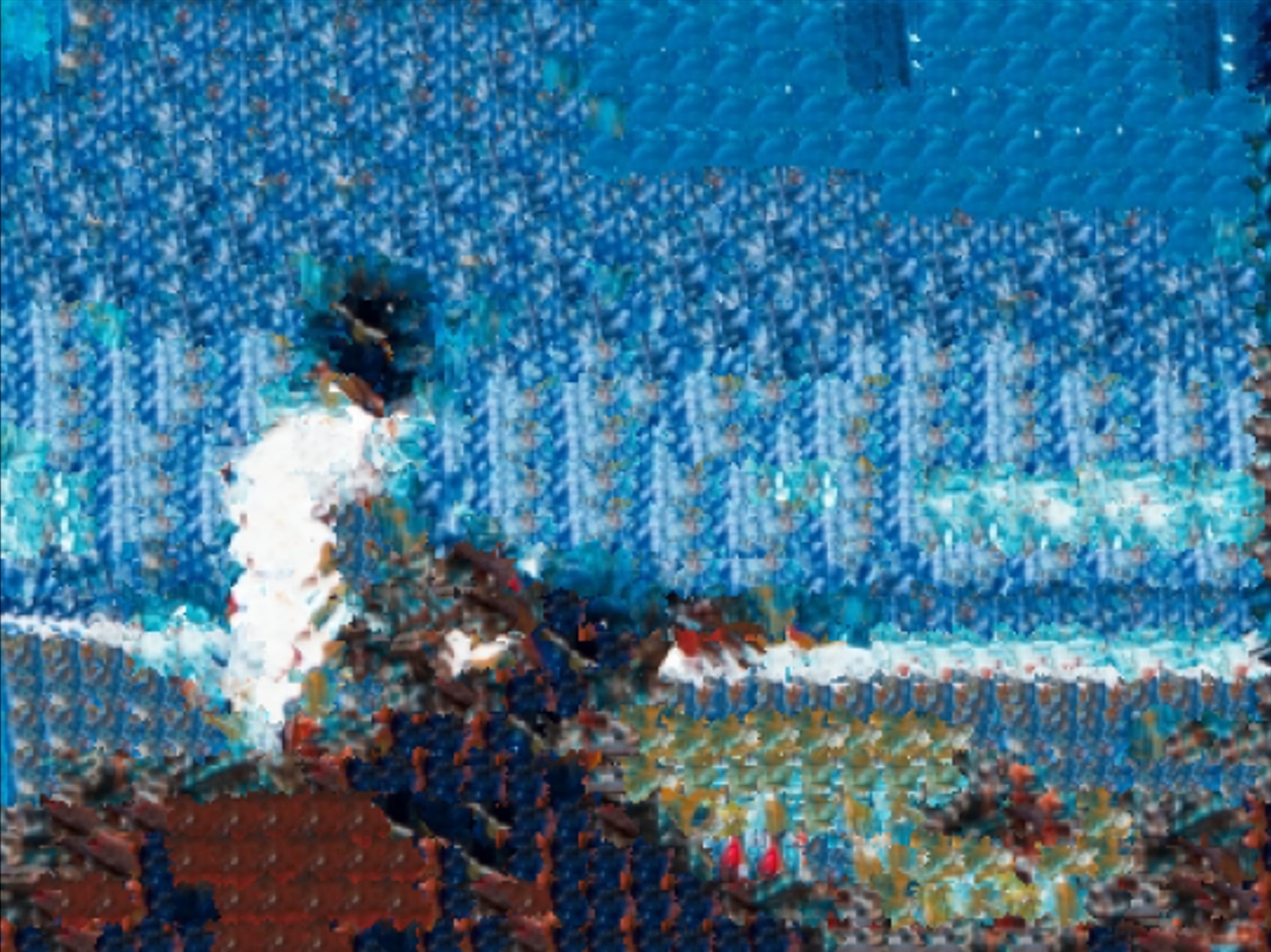}
  \end{subfigure}
  \setcounter{subfigure}{0}
  \hfill
  
  \begin{subfigure}[T]{0.15\textwidth}
     \includegraphics[width=1\columnwidth, height=0.75\columnwidth]{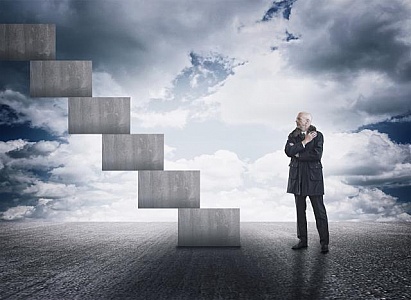}
     \caption{Input Image}
  \end{subfigure}
  ~
  \begin{subfigure}[T]{0.15\textwidth}
     \includegraphics[width=1\columnwidth, height=0.75\columnwidth]{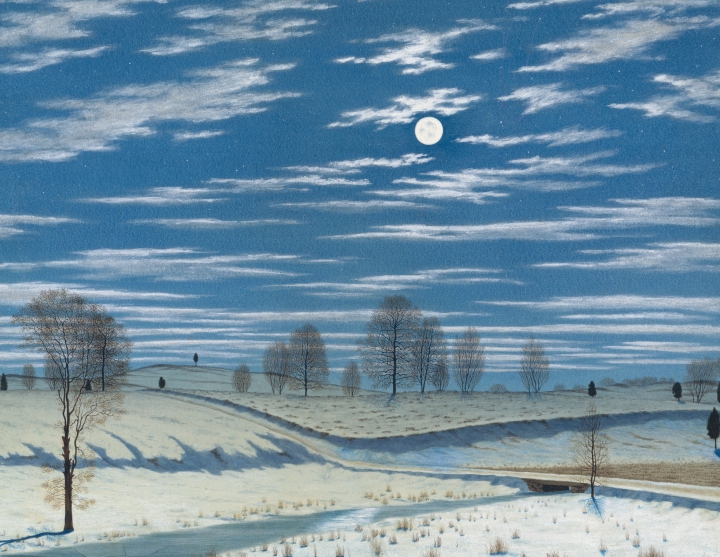}
     \caption{Style Image}
  \end{subfigure}
  ~
  \begin{subfigure}[T]{0.15\textwidth}
     \includegraphics[width=1\columnwidth, height=0.75\columnwidth]{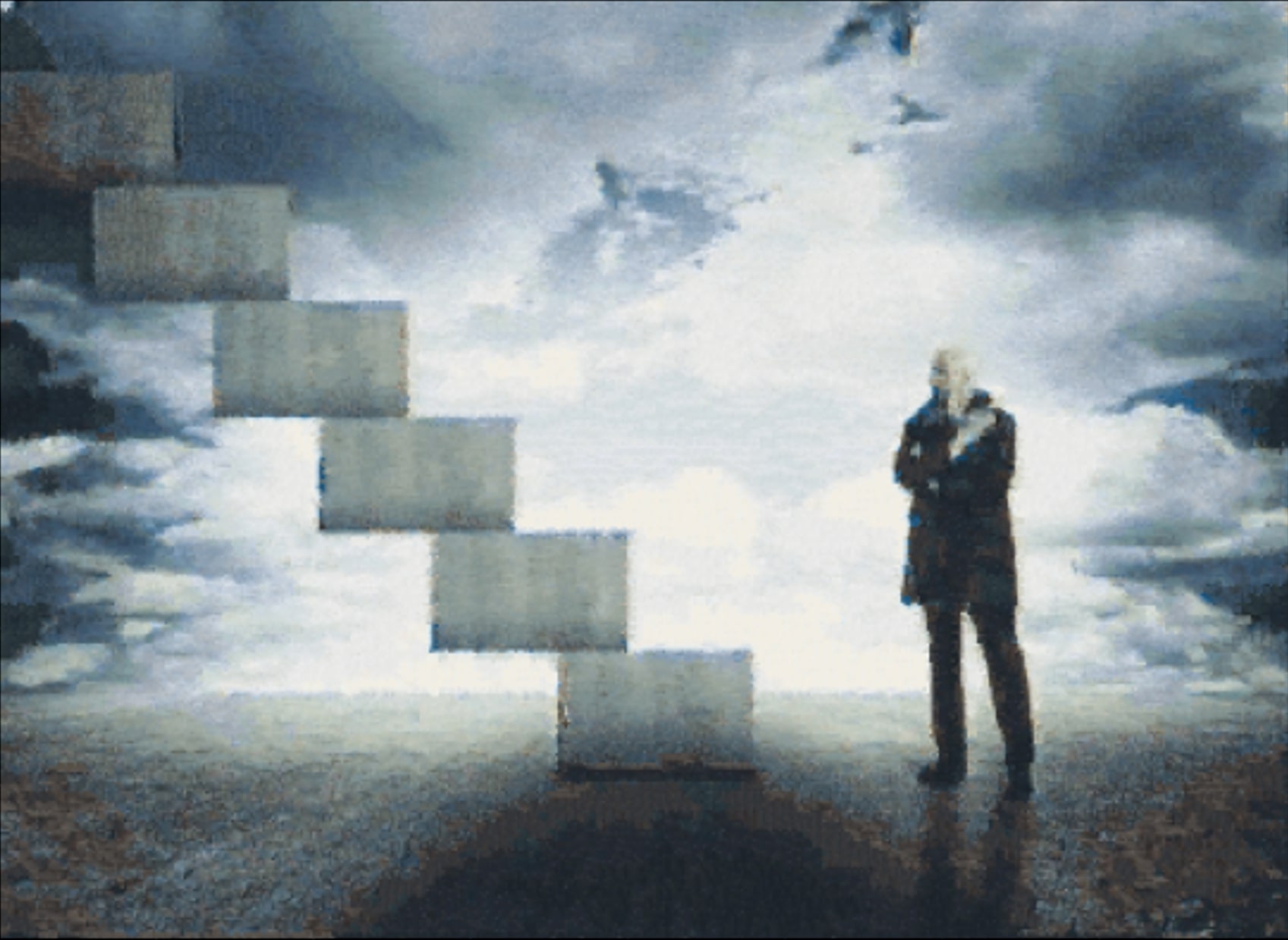}
     \caption{Generated Image with Patch Size as 5}
  \end{subfigure}
  ~
  \begin{subfigure}[T]{0.15\textwidth}
     \includegraphics[width=1\columnwidth, height=0.75\columnwidth]{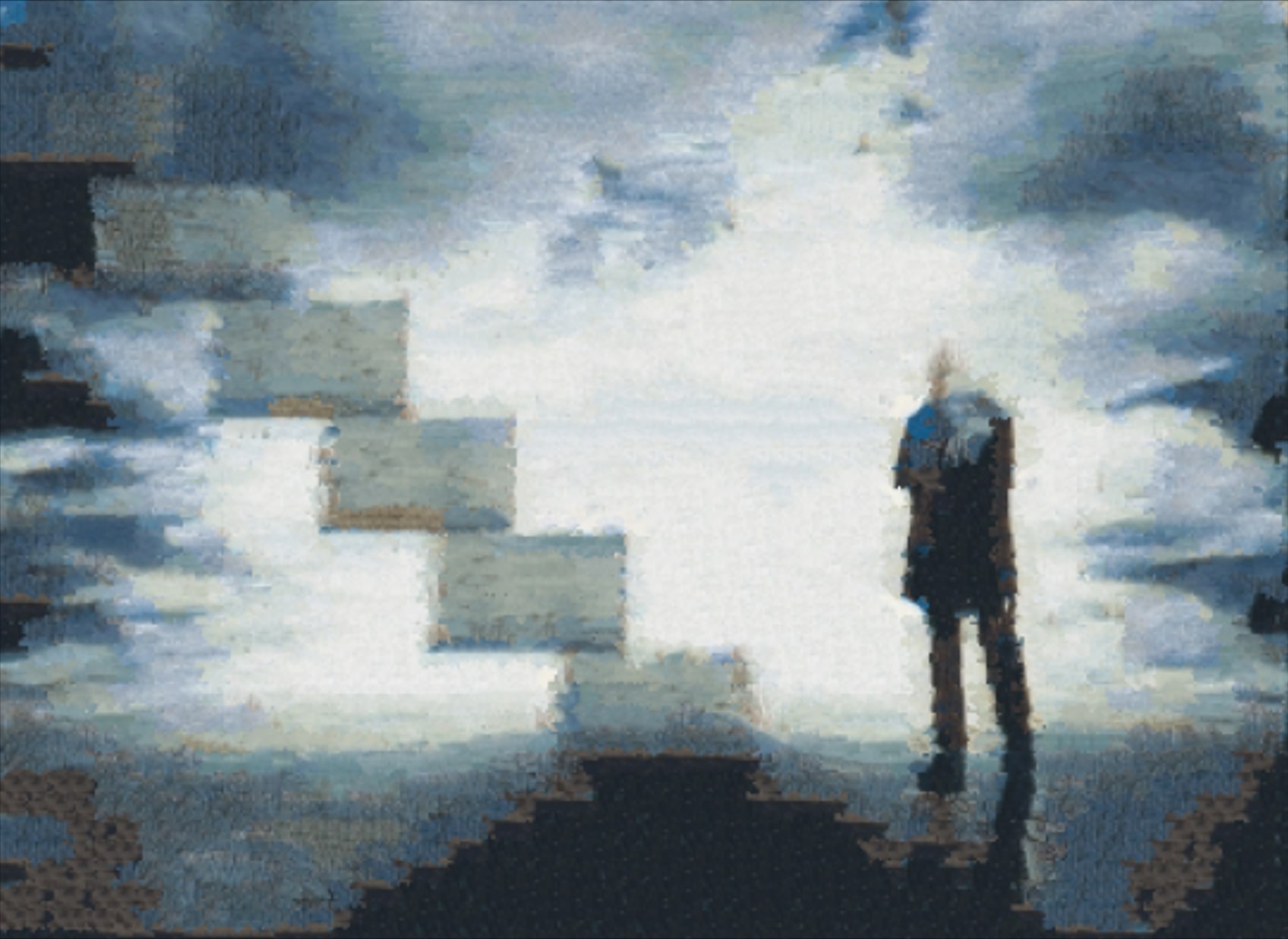}
     \caption{Generated Image with Patch Size as 11}
  \end{subfigure}
  ~
  \begin{subfigure}[T]{0.15\textwidth}
     \includegraphics[width=1\columnwidth, height=0.75\columnwidth]{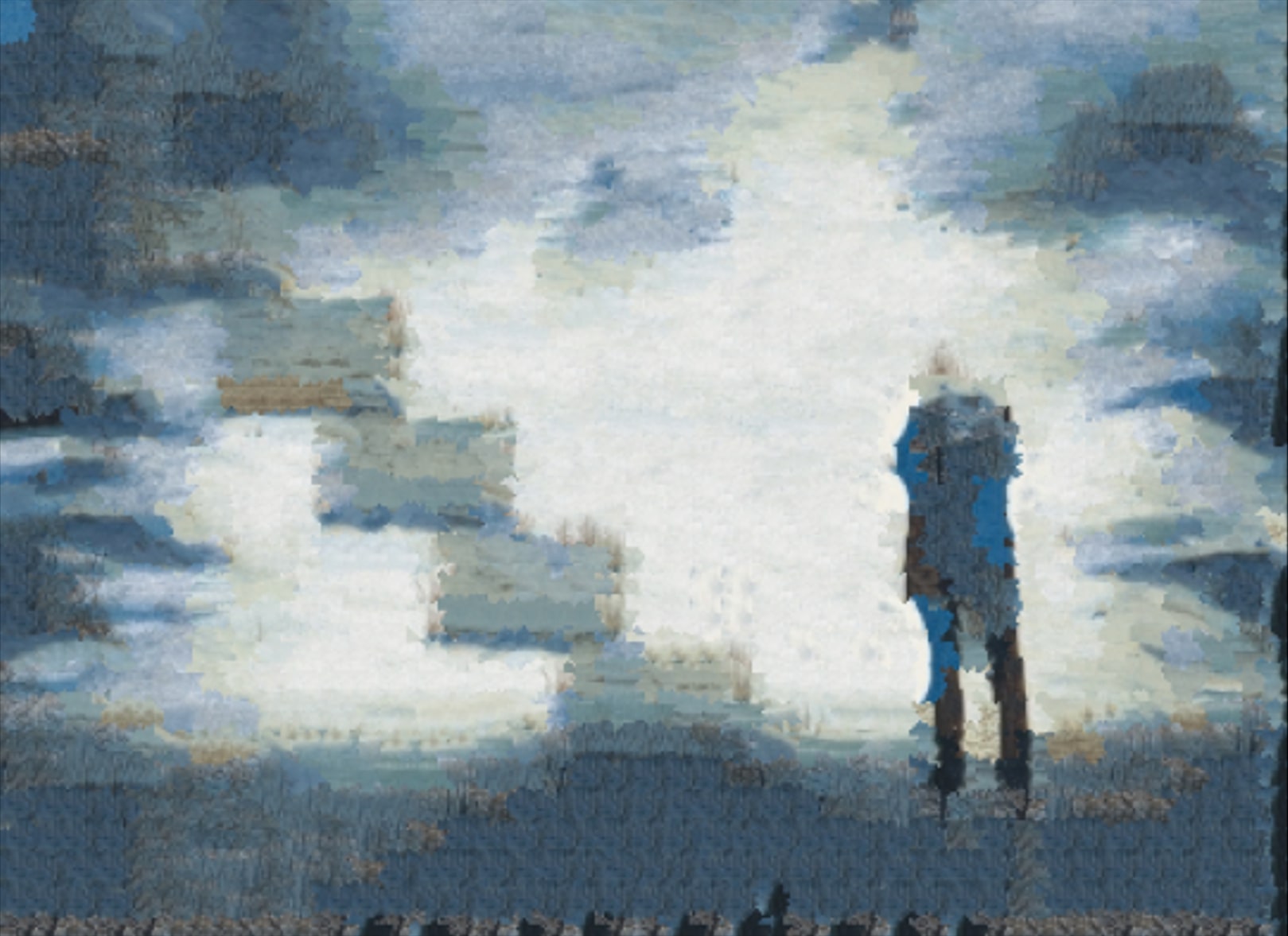}
     \caption{Generated Image with Patch Size as 16}
  \end{subfigure}
  ~
  \begin{subfigure}[T]{0.15\textwidth}
     \includegraphics[width=1\columnwidth, height=0.75\columnwidth]{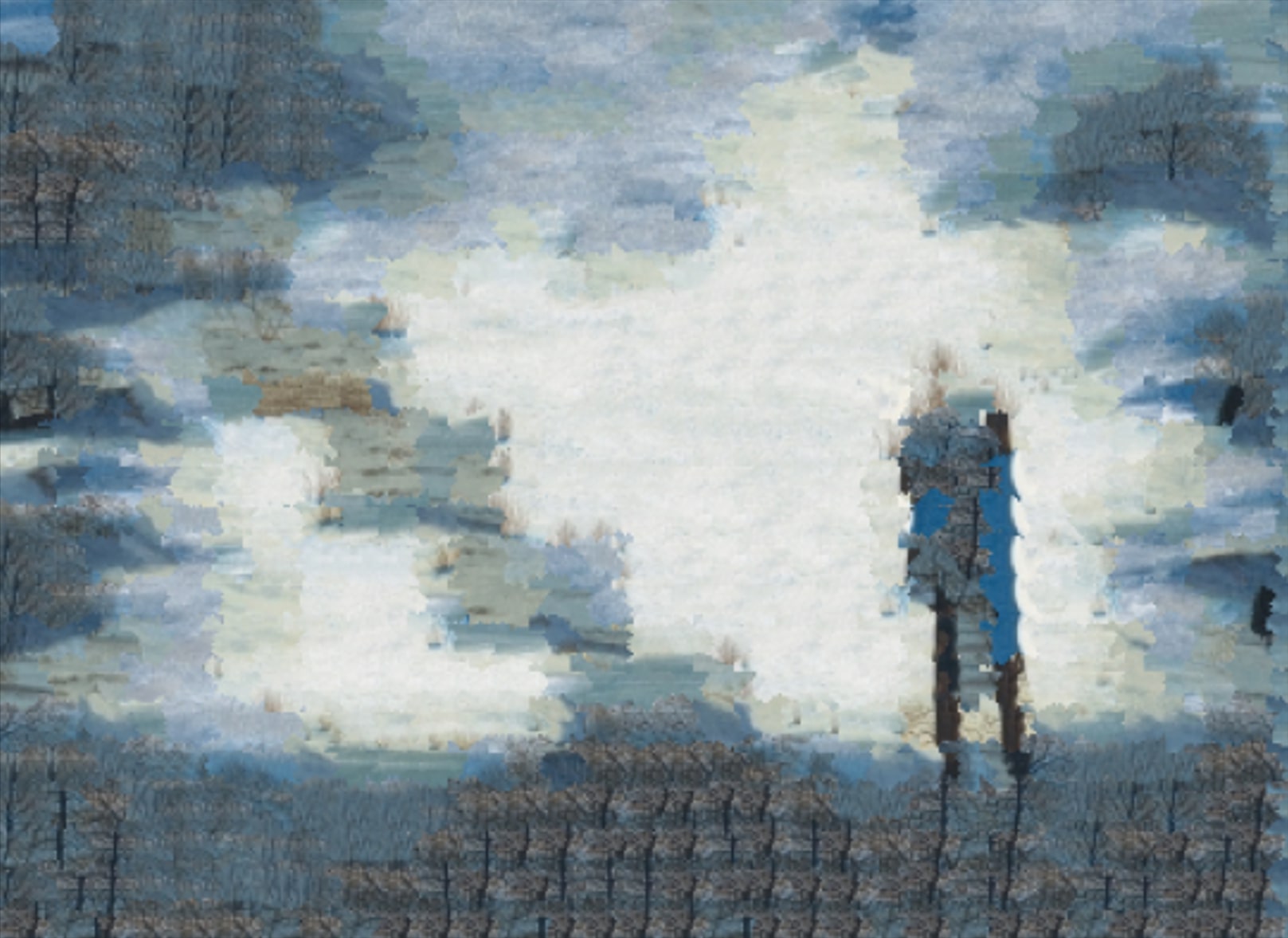}
     \caption{Generated Image with Patch Size as 20}
  \end{subfigure}
  \caption{Illustrates the extensive analysis of the traditional style transfer method by Alexei A. Efros et al. \cite{textureTransfer} on more complicated figures with more complicated styles. The patch size of the algorithm is set up as 5, 11, 16, and 20 respectively.}
  \label{fig:extend-study}
\end{figure*}

However, the effectiveness of the method on more complex styles hasn't been verified. In real world, style can display in more complicated and more various patterns. In addition, the input images could have more space information such as foreground objects and background landscape. Under such circumstances, the performance of such patch-splicing method is questionable. Therefore, to further explore the method's performance under more kinds of cases, we select the PASCAL VOC 2012 dataset \cite{engstrom2016faststyletransfer} containing diverse images. We also tested the algorithm with different patch sizes to study the performance when the patch size is set up differently.

As illustrated in Fig. \ref{fig:extend-study}, we selected a variety of input images paired with a diverse range of style images to evaluate the algorithm’s performance. However, the overall results are less than ideal, particularly given the complexity of both the style and input images. On one hand, the algorithm struggles to effectively capture the intricate style patterns. While increasing the patch size improves the retention of color styles, it adversely affects the fidelity of line patterns. Conversely, when the patch size is relatively small, the generated image resembles a pixel-level replacement rather than a true style transfer. On the other hand, the algorithm also fails to fully utilize the structural information inherent in the input images, resulting in a loss of detail and coherence. This suggests that the algorithm lacks the capability to perform effective style transfer on input images with complex information when provided with style images featuring intricate patterns. 

\subsection{Comparative Study}
To compare the the performance of the representative traditional style transfer method \cite{textureTransfer} and the representative deep-learning-based style transfer method \cite{ding2024style}, we continuously test both methods on the PASCAL VOC 2012 dataset \cite{engstrom2016faststyletransfer}. 

\begin{figure}[h!]
  \centering
  \begin{subfigure}[T]{0.24\linewidth}
    \includegraphics[width=\linewidth, height=0.75\linewidth]{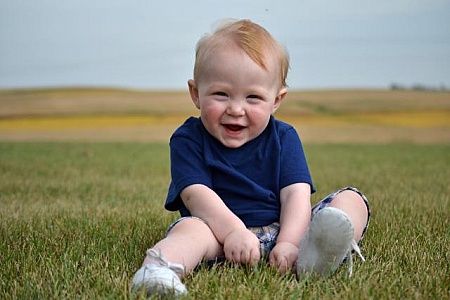}
  \end{subfigure}
  \begin{subfigure}[T]{0.24\linewidth}
    \includegraphics[width=\linewidth, height=0.75\linewidth]{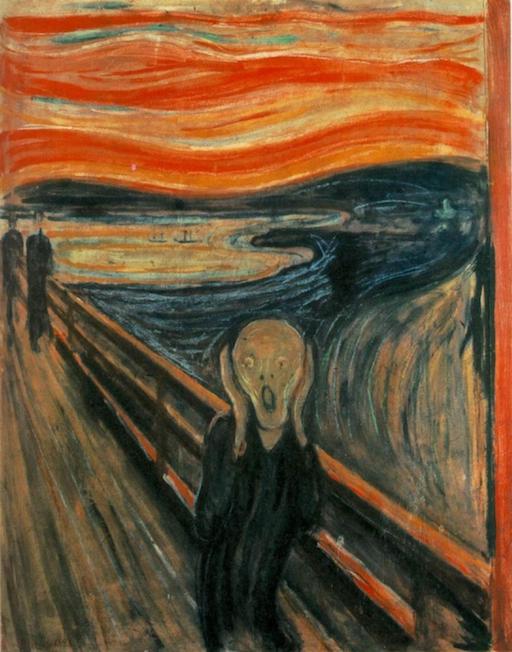}
  \end{subfigure}
  \begin{subfigure}[T]{0.24\linewidth}
    \includegraphics[width=\linewidth, height=0.75\linewidth]{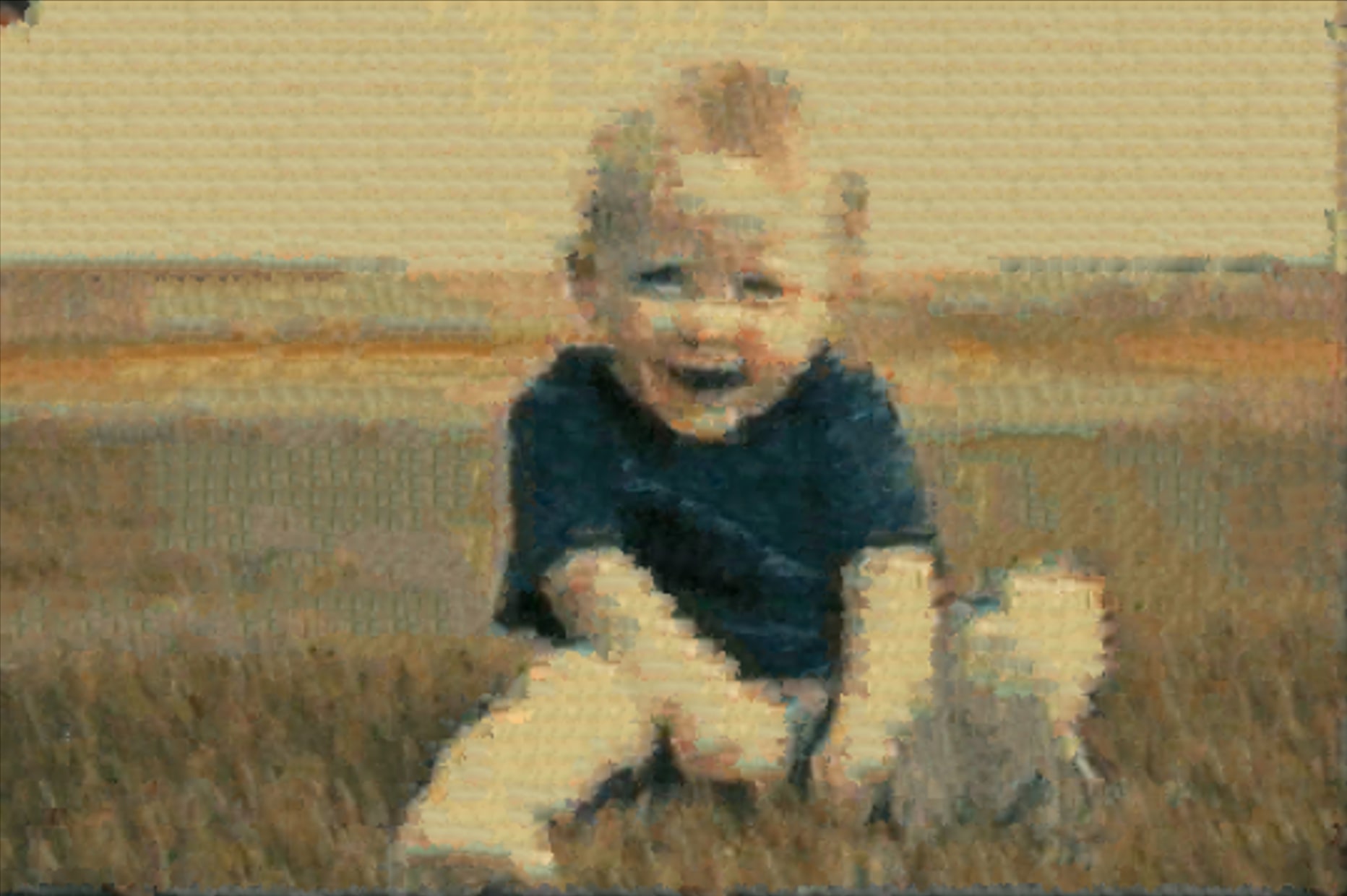}
  \end{subfigure}
  \begin{subfigure}[T]{0.24\linewidth}
    \includegraphics[width=\linewidth, height=0.75\linewidth]{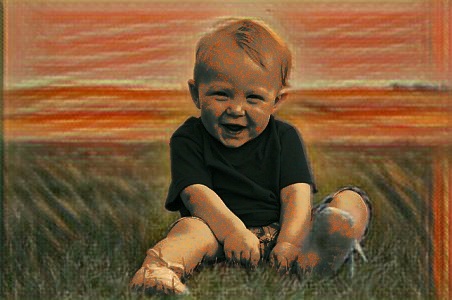}
  \end{subfigure}
  
  \centering
  \begin{subfigure}[T]{0.24\linewidth}
    \includegraphics[width=\linewidth, height=0.75\linewidth]{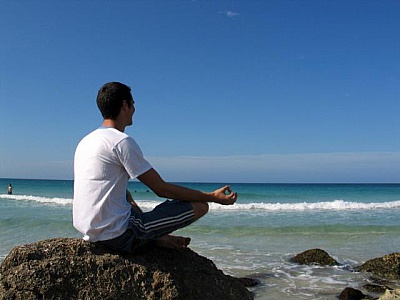}
  \end{subfigure}
  \begin{subfigure}[T]{0.24\linewidth}
    \includegraphics[width=\linewidth, height=0.75\linewidth]{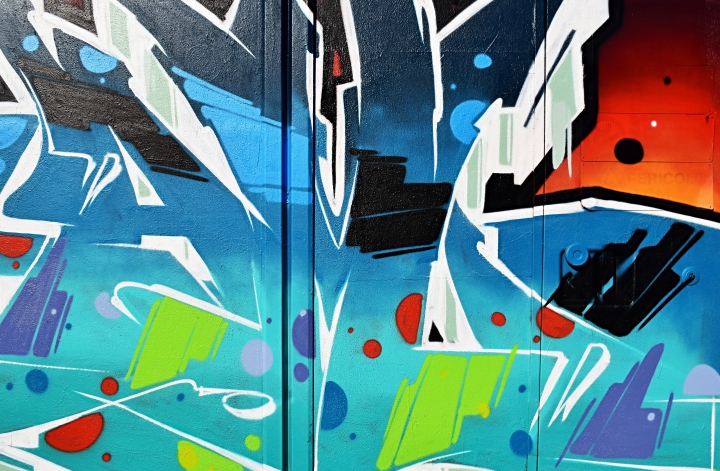}
  \end{subfigure}
  \begin{subfigure}[T]{0.24\linewidth}
    \includegraphics[width=\linewidth, height=0.75\linewidth]{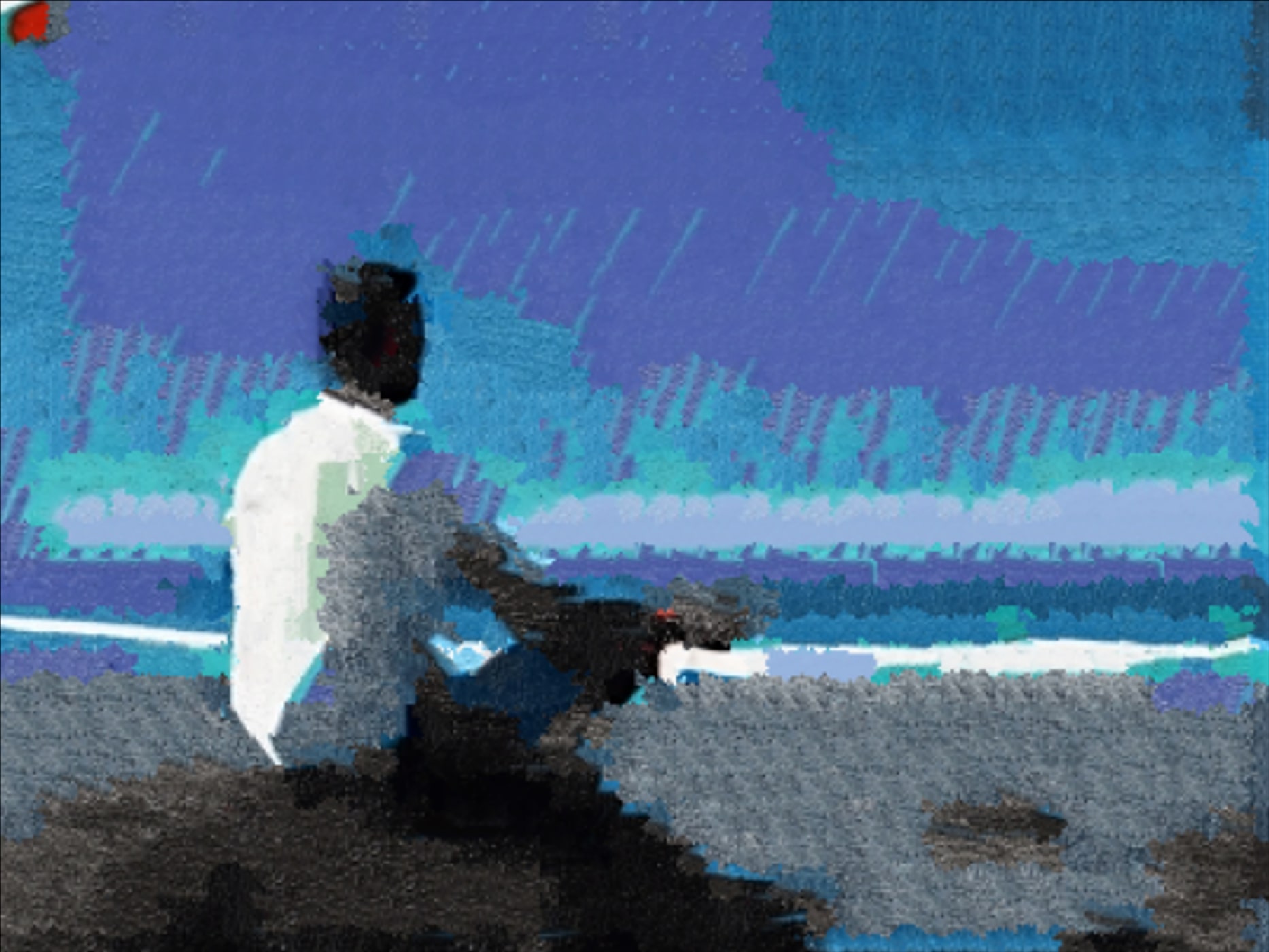}
  \end{subfigure}
  \begin{subfigure}[T]{0.24\linewidth}
    \includegraphics[width=\linewidth, height=0.75\linewidth]{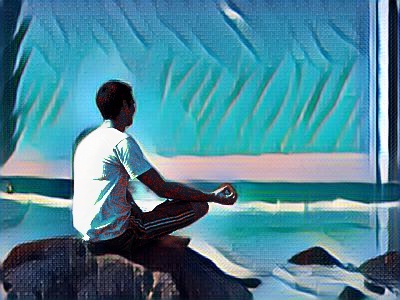}
  \end{subfigure}
  
  \centering
  \begin{subfigure}[T]{0.24\linewidth}
    \includegraphics[width=\linewidth, height=0.75\linewidth]{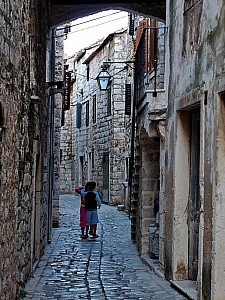}
  \end{subfigure}
  \begin{subfigure}[T]{0.24\linewidth}
    \includegraphics[width=\linewidth, height=0.75\linewidth]{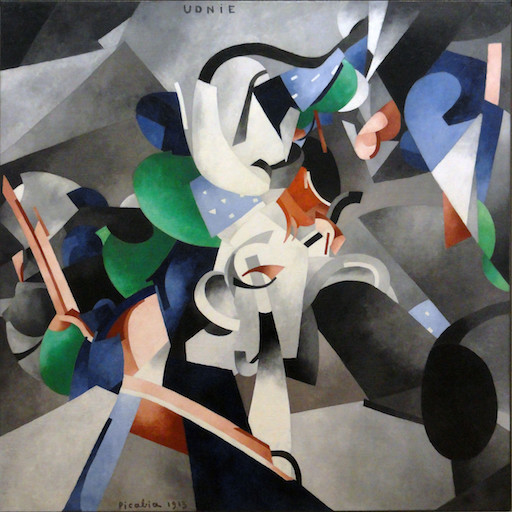}
  \end{subfigure}
  \begin{subfigure}[T]{0.24\linewidth}
    \includegraphics[width=\linewidth, height=0.75\linewidth]{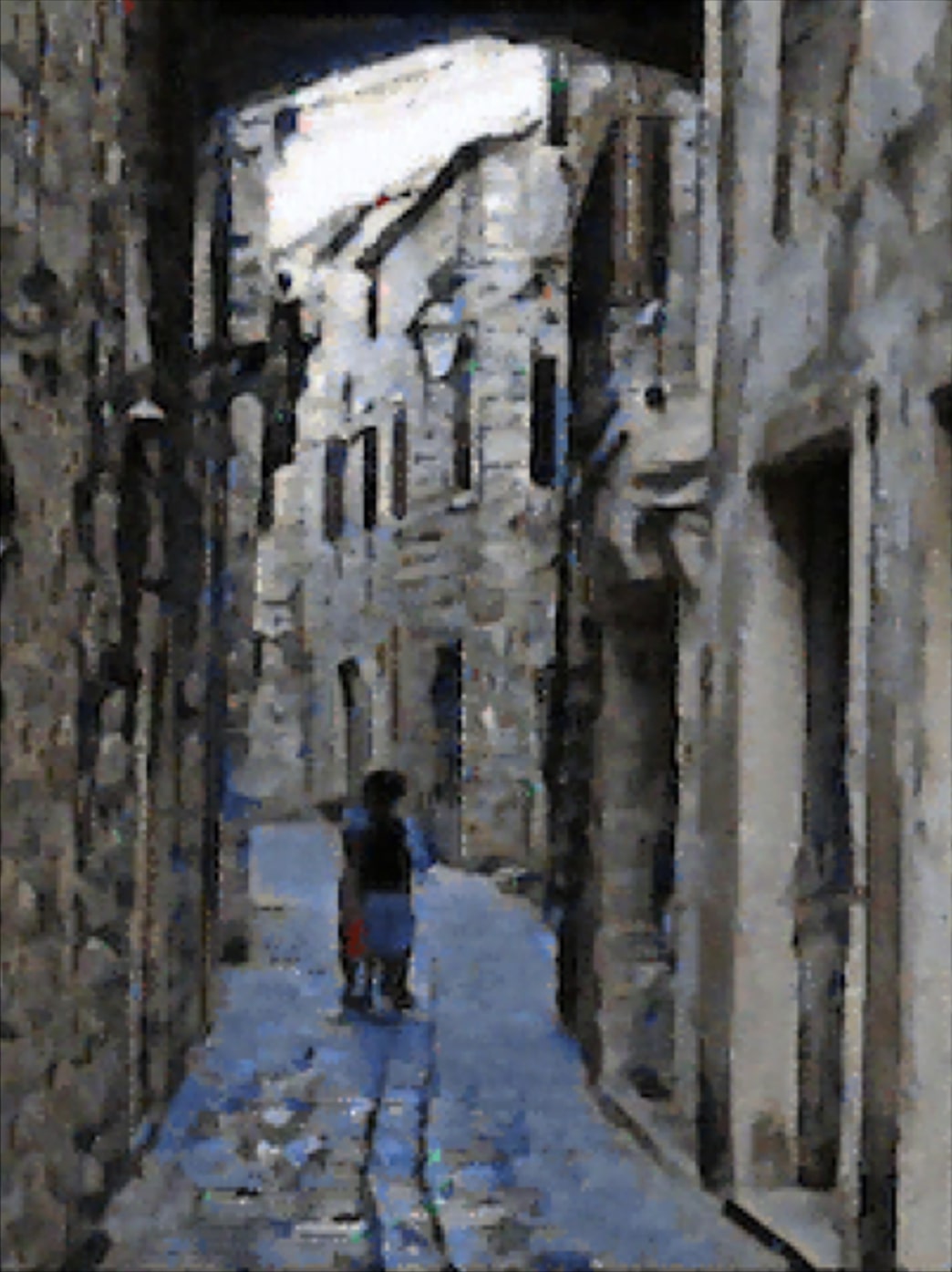}
  \end{subfigure}
  \begin{subfigure}[T]{0.24\linewidth}
    \includegraphics[width=\linewidth, height=0.75\linewidth]{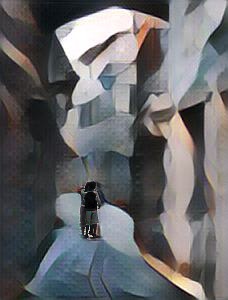}
  \end{subfigure}

  \begin{subfigure}[T]{0.24\linewidth}
    \includegraphics[width=\linewidth, height=0.75\linewidth]{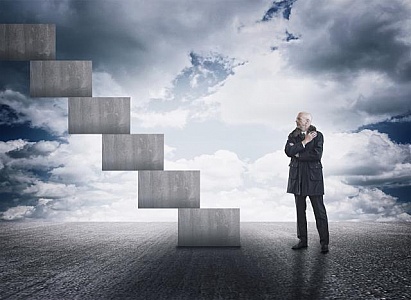}
    \caption{}
  \end{subfigure}
  \begin{subfigure}[T]{0.24\linewidth}
    \includegraphics[width=\linewidth, height=0.75\linewidth]{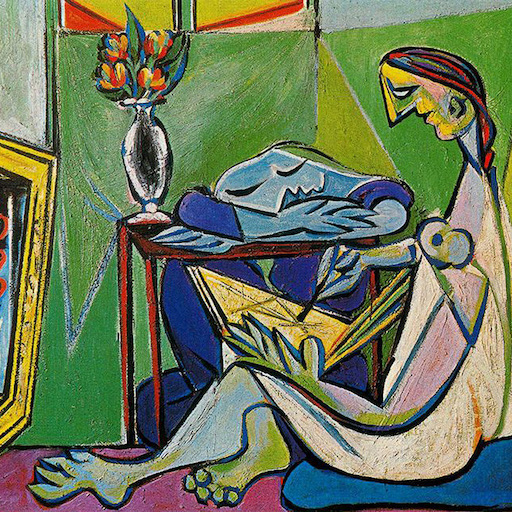}
    \caption{}
  \end{subfigure}
  \begin{subfigure}[T]{0.24\linewidth}
  \includegraphics[width=\linewidth, height=0.75\linewidth]{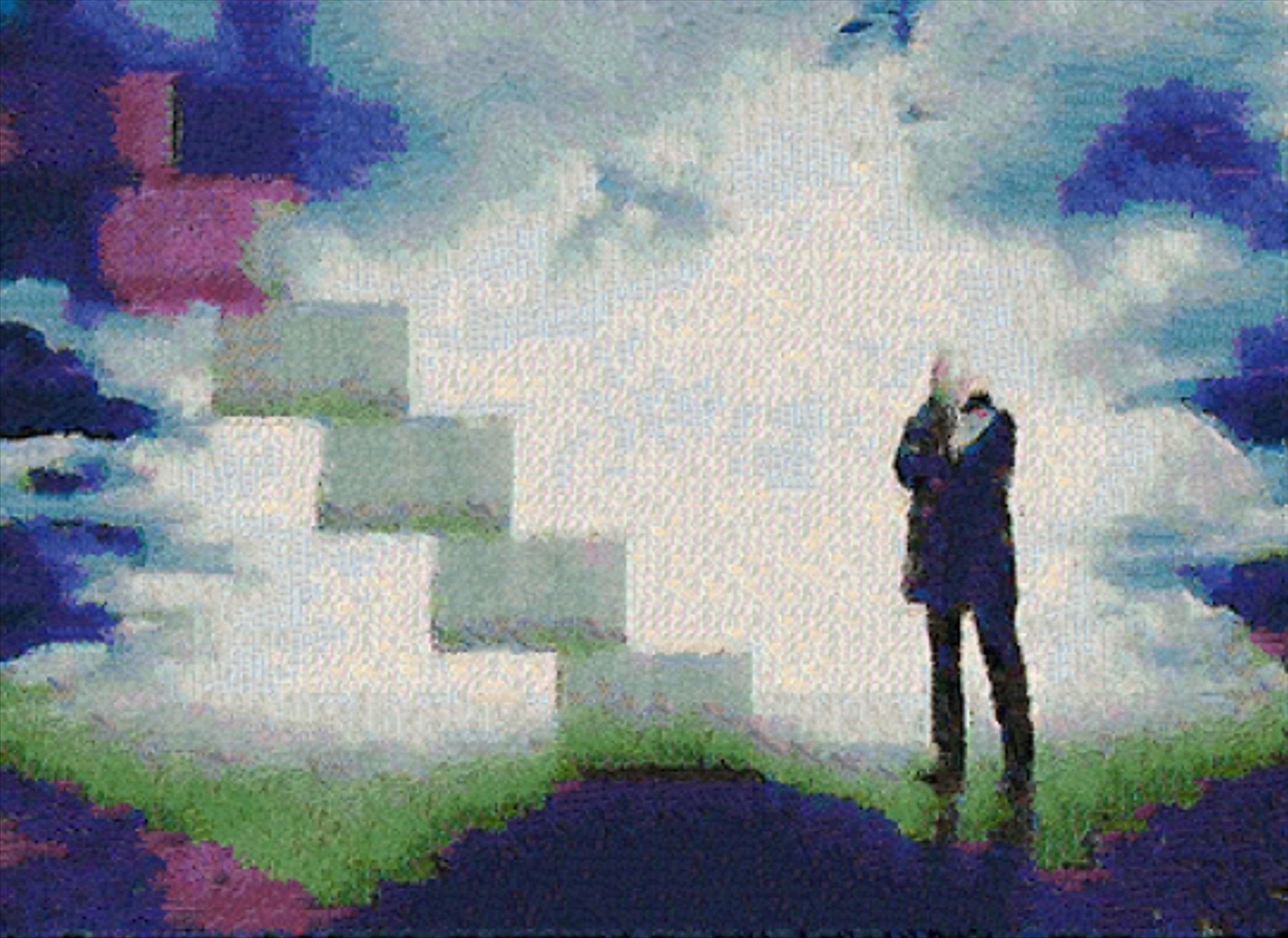}
    \caption{}
  \end{subfigure}
  \begin{subfigure}[T]{0.24\linewidth}
  \includegraphics[width=\linewidth, height=0.75\linewidth]{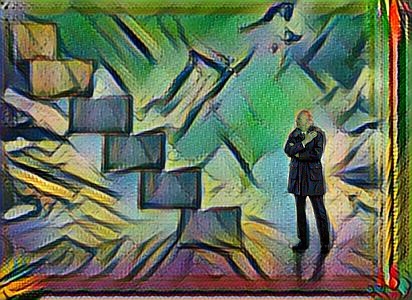}
    \caption{}
  \end{subfigure}
  \caption{Comparison between the representative traditional style transfer method \cite{textureTransfer} and the representative style transfer method \cite{ding2024style}. (a) Input image. (b) Style image. (c) Generated image by the representative traditional style transfer method. (d) Generated by the representative deep-learning-based style transfer method.}
  \label{fig:cmp-study}
\end{figure}

As shown in Fig. \ref{fig:cmp-study}, the generated images from both methods exhibit significant differences. From the perspective of color, the deep-learning-based method transfers the image style in a smoother and more vivid manner. This is clearly reflected in the generated results of the first, second, and fourth rows. The deep-learning-based images display a richer and more diverse color combination. In contrast, the images produced by the traditional method exhibit poorer color selection, with some common colors from the style image barely visible in the generated images (e.g., the color red is noticeably absent in the traditional method’s generated image in the first row). 

From the perspective of line and structural information, the traditional method, due to its patch placement nature, preserves the major structural information of the original input image. On the other hand, the deep-learning-based method slightly alters the spatial information, as seen in the generated images of the third row. However, this characteristic of the traditional method can also create unnatural picture chunks. For example, in the generated images of the second row, there are unnatural lines visible in the sky. As mentioned in the previous section, this issue is related to the patch size selection, which can be challenging to optimize in practice.

\section{Conclusions}
\subsection{Conclusions}
This paper presents a comprehensive comparative study between the representative traditional method \cite{textureTransfer} and the representative style transfer method \cite{ding2024style}. The analysis indicates that the traditional method struggles to capture the complexity of style patterns effectively. While testing different patch sizes, it was found that increasing the patch size improves the retention of color styles; however, it negatively impacts the fidelity of line patterns. Conversely, smaller patch sizes result in outputs that resemble pixel-level replacements rather than authentic style transfers. Additionally, the algorithm does not adequately utilize the structural information of the input images, leading to a loss of detail and coherence in the generated outputs. These findings highlight the traditional method’s shortcomings in performing style transfer when dealing with complex input images and intricate style references.

As for the comparative study with a deep-learning-based method, the researchers observed significant differences in the generated images from both approaches. The deep-learning method excels in color transfer, producing smoother and more vivid results with richer color combinations, as seen in the generated images of the first, second, and fourth rows. In contrast, the traditional method demonstrates a weaker color palette. From the perspective of line and structural information, the traditional method maintains the major structural elements of the original input image due to its patch placement strategy. However, this can result in unnatural visual artifacts, such as awkward lines in generated images. On the other hand, the deep-learning method slightly alters the spatial information, which can lead to inconsistencies in structure but ultimately provides a more cohesive overall aesthetic.

The study concludes that while the traditional patch-splicing method has certain strengths in preserving structure, it is inadequate for effective style transfer in complex scenarios. The deep-learning-based approach offers superior performance in color richness and overall visual quality, demonstrating a need for advanced techniques to address the limitations of traditional methods in style transfer applications. 

\subsection{Future Work}
While this study provides a thorough analysis comparing traditional and deep-learning-based style transfer methods, several limitations could serve as avenues for future research. First of all, the cost-effectiveness of both traditional and deep-learning-based methods deserves further examination. Understanding the computational and resource requirements of each approach will be crucial for evaluating their practicality in real-world applications. Secondly, additional studies incorporating a wider range of diverse strategies should be undertaken to enable a more comprehensive discussion of their strengths and weaknesses. Exploring hybrid techniques that combine elements from both paradigms could lead to innovative solutions that better capture artistic styles \cite{10636773}. 

\renewcommand{\bibfont}{\footnotesize}

\footnotesize{
\bibliographystyle{IEEEtran}
\bibliography{main}

\begin{thebibliography}{10}
\providecommand{\url}[1]{#1}
\csname url@samestyle\endcsname
\providecommand{\newblock}{\relax}
\providecommand{\bibinfo}[2]{#2}
\providecommand{\BIBentrySTDinterwordspacing}{\spaceskip=0pt\relax}
\providecommand{\BIBentryALTinterwordstretchfactor}{4}
\providecommand{\BIBentryALTinterwordspacing}{\spaceskip=\fontdimen2\font plus
\BIBentryALTinterwordstretchfactor\fontdimen3\font minus \fontdimen4\font\relax}
\providecommand{\BIBforeignlanguage}[2]{{%
\expandafter\ifx\csname l@#1\endcsname\relax
\typeout{** WARNING: IEEEtran.bst: No hyphenation pattern has been}%
\typeout{** loaded for the language `#1'. Using the pattern for}%
\typeout{** the default language instead.}%
\else
\language=\csname l@#1\endcsname
\fi
#2}}
\providecommand{\BIBdecl}{\relax}
\BIBdecl

\bibitem{xiao2024convolutional}
M.~Xiao, Y.~Li, X.~Yan, M.~Gao, and W.~Wang, ``Convolutional neural network classification of cancer cytopathology images: taking breast cancer as an example,'' \emph{arXiv preprint arXiv:2404.08279}, 2024.

\bibitem{jin-etal-2022-deep}
D.~Jin \emph{et~al.}, ``Deep learning for text style transfer: A survey,'' \emph{Computational Linguistics}, vol.~48, no.~1, pp. 155--205, Mar. 2022.

\bibitem{hu2024rapsppp}
W.~Hu, J.-B. Uwineza, and J.~A. Farrell, ``Outlier accommodation for gnss precise point positioning using risk-averse state estimation,'' \emph{arXiv preprint arXiv:2402.01860}, 2024.

\bibitem{ding2024confidence}
Z.~Ding \emph{et~al.}, ``Confidence trigger detection: Accelerating real-time tracking-by-detection systems,'' \emph{arXiv preprint arXiv:1902.00615}, 2024.

\bibitem{hu2024optimization}
W.~Hu, Y.~Hu, M.~Stas, and J.~A. Farrell, ``Optimization-based outlier accommodation for tightly coupled rtk-aided inertial navigation systems in urban environments,'' \emph{arXiv preprint arXiv:2407.13912}, 2024.

\bibitem{li2019segmentation}
P.~Li \emph{et~al.}, ``Contextual hourglass network for semantic segmentation of high resolution aerial imagery,'' \emph{arXiv preprint arXiv:1810.12813}, 2019.

\bibitem{yan2024survival}
X.~Yan \emph{et~al.}, ``Survival prediction across diverse cancer types using neural networks,'' \emph{arXiv preprint arXiv:2404.08713}, 2024.

\bibitem{ding2024llava}
Z.~Ding \emph{et~al.}, ``Enhance image-to-image generation with llava prompt and negative prompt,'' \emph{arXiv preprint arXiv:2406.01956}, 2024.

\bibitem{gatys2015neural}
L.~A. Gatys \emph{et~al.}, ``A neural algorithm of artistic style,'' \emph{arXiv preprint 1508.06576}, 2024.

\bibitem{textureTransfer}
A.~A. Efros and W.~T. Freeman, ``Image quilting for texture synthesis and transfer,'' in \emph{SIGGRAPH}, 2001, p. 341–346.

\bibitem{li2023deception}
P.~Li \emph{et~al.}, ``Deception detection from linguistic and physiological data streams using bimodal convolutional neural networks,'' \emph{arXiv preprint arXiv:2311.10944}, 2023.

\bibitem{zhou2024reconstruction}
Y.~Zhou \emph{et~al.}, ``Evaluating modern approaches in 3d scene reconstruction: Nerf vs gaussian-based methods,'' \emph{arXiv preprint arXiv:2408.04268}, 2024.

\bibitem{ni2024earnings}
H.~Ni \emph{et~al.}, ``Harnessing earnings reports for stock predictions: A qlora-enhanced llm approach,'' \emph{arXiv preprint arXiv:2408.06634}, 2024.

\bibitem{hu22usingppp}
W.~Hu, A.~Neupane, and J.~A. Farrell, ``Using ppp information to implement a global real-time virtual network dgnss approach,'' \emph{IEEE Trans. Veh. Technol.}, vol.~71, no.~10, pp. 10\,337--10\,349, 2022.

\bibitem{alzubaidi2021review}
L.~Alzubaidi \emph{et~al.}, ``Review of deep learning: concepts, cnn architectures, challenges, applications, future directions,'' \emph{Journal of big Data}, vol.~8, pp. 1--74, 2021.

\bibitem{tong2022normal}
W.~Tong \emph{et~al.}, ``Normal assisted pixel-visibility learning with cost aggregation for multiview stereo,'' \emph{IEEE Transactions on Intelligent Transportation Systems}, vol.~23, no.~12, pp. 24\,686--24\,697, 2022.

\bibitem{li2024vqa}
P.~Li \emph{et~al.}, ``Exploring diverse methods in visual question answering,'' \emph{arXiv preprint arXiv:2404.13565}, 2024.

\bibitem{li2023investigation}
Y.~Li \emph{et~al.}, ``Investigation of creating accessibility linked data based on publicly available accessibility datasets,'' in \emph{ICCNS}, 2023, pp. 77--81.

\bibitem{ni2024timeseries}
H.~Ni \emph{et~al.}, ``Time series modeling for heart rate prediction: From arima to transformers,'' \emph{arXiv preprint arXiv:2406.12199}, 2024.

\bibitem{wuinversionnet}
Y.~Wu \emph{et~al.}, ``Inversionnet-accurate and efficient data-driven seismic waveform-inversion with convolutional neural networks,'' \emph{ResearchGate}, 2018.

\bibitem{ZHANG2024103951}
Z.~Zhang \emph{et~al.}, ``Rethink arbitrary style transfer with transformer and contrastive learning,'' \emph{Computer Vision and Image Understanding}, vol. 241, p. 103951, 2024.

\bibitem{zheng2024advanced}
Q.~Zheng \emph{et~al.}, ``Advanced payment security system:xgboost, lightgbm and smote integrated,'' \emph{arXiv preprint arXiv:2406.04658}, 2024.

\bibitem{o2015introduction}
K.~O'shea and R.~Nash, ``An introduction to convolutional neural networks,'' \emph{arXiv preprint arXiv:1511.08458}, 2015.

\bibitem{202407.0981}
X.~Li \emph{et~al.}, ``A vehicle classification method based on machine learning,'' \emph{Preprints}, July 2024.

\bibitem{dai23}
W.~Dai \emph{et~al.}, ``Addressing unintended bias in toxicity detection: An lstm and attention-based approach,'' in \emph{ICAICA}, 2023, pp. 375--379.

\bibitem{simonyan2014very}
K.~Simonyan and A.~Zisserman, ``Very deep convolutional networks for large-scale image recognition,'' \emph{arXiv preprint arXiv:1409.1556}, 2014.

\bibitem{202407.2102}
X.~Li \emph{et~al.}, ``Intelligent vehicle classification system based on deep learning and multi-sensor fusion,'' \emph{Preprints}, July 2024.

\bibitem{bian2024diffusion}
W.~Bian \emph{et~al.}, ``Diffusion modeling with domain-conditioned prior guidance for accelerated mri and qmri reconstruction,'' \emph{IEEE Transactions on Medical Imaging}, 2024.

\bibitem{wang2016unsupervised}
T.~Y. Wang \emph{et~al.}, ``Unsupervised texture transfer from images to model collections.'' \emph{ACM Trans. Graph.}, vol.~35, no.~6, pp. 177--1, 2016.

\bibitem{bian2021optimization}
W.~Bian, Y.~Chen, X.~Ye, and Q.~Zhang, ``An optimization-based meta-learning model for mri reconstruction with diverse dataset,'' \emph{Journal of Imaging}, vol.~7, no.~11, p. 231, 2021.

\bibitem{bergen1988early}
J.~R. Bergen and E.~H. Adelson, ``Early vision and texture perception,'' \emph{Nature}, vol. 333, no. 6171, pp. 363--364, 1988.

\bibitem{jiashu2021performance}
H.~Jiashu, ``Performance analysis of facial recognition: A critical review through glass factor,'' in \emph{CDS}.\hskip 1em plus 0.5em minus 0.4em\relax IEEE, 2021, pp. 429--436.

\bibitem{bian2024review}
W.~Bian, ``A review of electromagnetic elimination methods for low-field portable mri scanner,'' \emph{arXiv preprint arXiv:2406.17804}, 2024.

\bibitem{Hoamiao2023}
N.~Chakraborty \emph{et~al.}, ``Beast: Building an embodied action-prediction system with trajectory data,'' in \emph{Alexa Prize SimBot Challenge Proceedings}, 2023.

\bibitem{DBLP:conf/iclr/XuYLC24}
H.~Xu \emph{et~al.}, ``Can speculative sampling accelerate react without compromising reasoning quality?'' in \emph{Tiny Papers @ {ICLR} 2024}, 2024.

\bibitem{hly}
L.~He \emph{et~al.}, ``Cognitive models for abacus gesture learning,'' \emph{Proceedings of the Annual Meeting of the Cognitive Science Society}, vol.~46, 2024.

\bibitem{ding2024style}
Z.~Ding, P.~Li, Q.~Yang, S.~Li, and Q.~Gong, ``Regional style and color transfer,'' in \emph{CVIDL}.\hskip 1em plus 0.5em minus 0.4em\relax IEEE, 2024, pp. 593--597.

\bibitem{engstrom2016faststyletransfer}
L.~Engstrom, ``Fast style transfer,'' \emph{TensorFlow Hub}, 2024.

\bibitem{10636773}
Z.~Wang \emph{et~al.}, ``Cdc-yolofusion: Leveraging cross-scale dynamic convolution fusion for visible-infrared object detection,'' \emph{IEEE Transactions on Intelligent Vehicles}, pp. 1--14, 2024.

\end{thebibliography}
}

\end{document}